\documentclass{article}

\PassOptionsToPackage{numbers, compress}{natbib}
\usepackage[preprint]{neurips_2021_ml4ad}

\usepackage[utf8]{inputenc} % allow utf-8 input
\usepackage[T1]{fontenc}    % use 8-bit T1 fonts
\usepackage{hyperref}       % hyperlinks
\usepackage{url}            % simple URL typesetting
\usepackage{booktabs}       % professional-quality tables
\usepackage{amsfonts}       % blackboard math symbols
\usepackage{nicefrac}       % compact symbols for 1/2, etc.
\usepackage{microtype}      % microtypography
\usepackage{xcolor}         % colors
\usepackage{cite}
\usepackage{stfloats}
\usepackage{graphicx}
\usepackage{tikz}
\usepackage{subcaption}

\title{PolyTrack: Tracking with Bounding Polygons}

\author{%
  Gaspar Faure\\
   Polyechnique Montréal\\
  \texttt{gaspar.faure@polymtl.ca} \\
  \And Hughes Perreault \\
   Polyechnique Montréal\\
  \texttt{hughes.perreault@polymtl.ca} \\  
  \And Guillaume-Alexandre Bilodeau\\
     Polyechnique Montréal\\
  \texttt{gabilodeau@polymtl.ca} \\
  \And Nicolas Saunier\\
   Polyechnique Montréal\\
  \texttt{nicolas.saunier@polymtl.ca} 
}

\begin{document}

\maketitle

\begin{abstract}
In this paper, we present a novel method called PolyTrack for fast multi-object tracking and segmentation using bounding polygons. Polytrack detects objects by producing heatmaps of their center keypoint. For each of them, a rough segmentation is done by computing a bounding polygon over each instance instead of the traditional bounding box. Tracking is done by taking two consecutive frames as input and computing a center offset for each object detected in the first frame to predict its location in the second frame. A Kalman filter is also applied to reduce the number of ID switches. Since our target application is automated driving systems, we apply our method on urban environment videos. We trained and evaluated PolyTrack on the MOTS and KITTIMOTS datasets. Results show that tracking polygons can be a good alternative to bounding box and mask tracking. The code of PolyTrack is available at \url{https://github.com/gafaua/PolyTrack}.
\end{abstract}

\section{Introduction}

Multi-object tracking and segmentation (MOTS) is a fairly novel task that combines instance segmentation and multi-object tracking. It consists of associating and segmenting instances of objects corresponding to predetermined classes across multiple consecutive frames. This problem is important for many applications, in particular for intelligent transportation systems (ITS) and driver assistance and automation technologies. 

The MOTS task is quite complex as it requires multiple outputs across multiple frames. Additionally, producing segmentation masks can be slow and very demanding in memory, which is not suitable for many ITS applications that are either running on on-board computers or in the cloud. For this reason, our method was designed using fast and lightweight elements. PolyTrack is based on the real-time instance segmentation method CenterPoly~\citep{perreault2021centerpoly} and on the tracking module of the CenterTrack~\citep{zhou2020tracking} multi-object tracker. It is crucial to reduce the representation size of the output as well as the computational cost of producing it. CenterPoly uses polygons for the masks which are well suited for this context, taking up less space than full segmentation masks and being fast to produce.   

\begin{figure}[h]
\centering
\begin{subfigure}{.5\textwidth}
  \centering
  \includegraphics[width=0.8\linewidth]{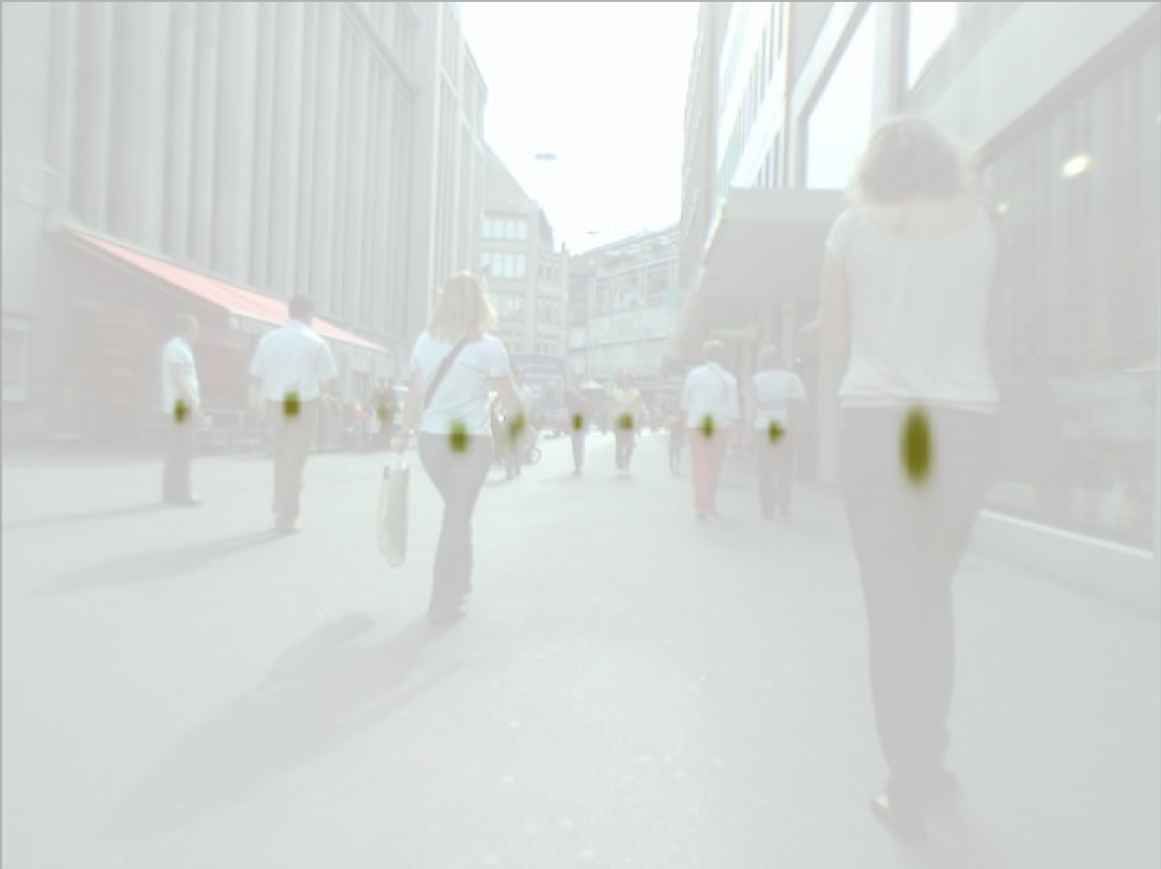}
  \caption{}
  \label{fig:sub1}
\end{subfigure}%
\begin{subfigure}{.5\textwidth}
  \centering
  \includegraphics[width=0.8\linewidth]{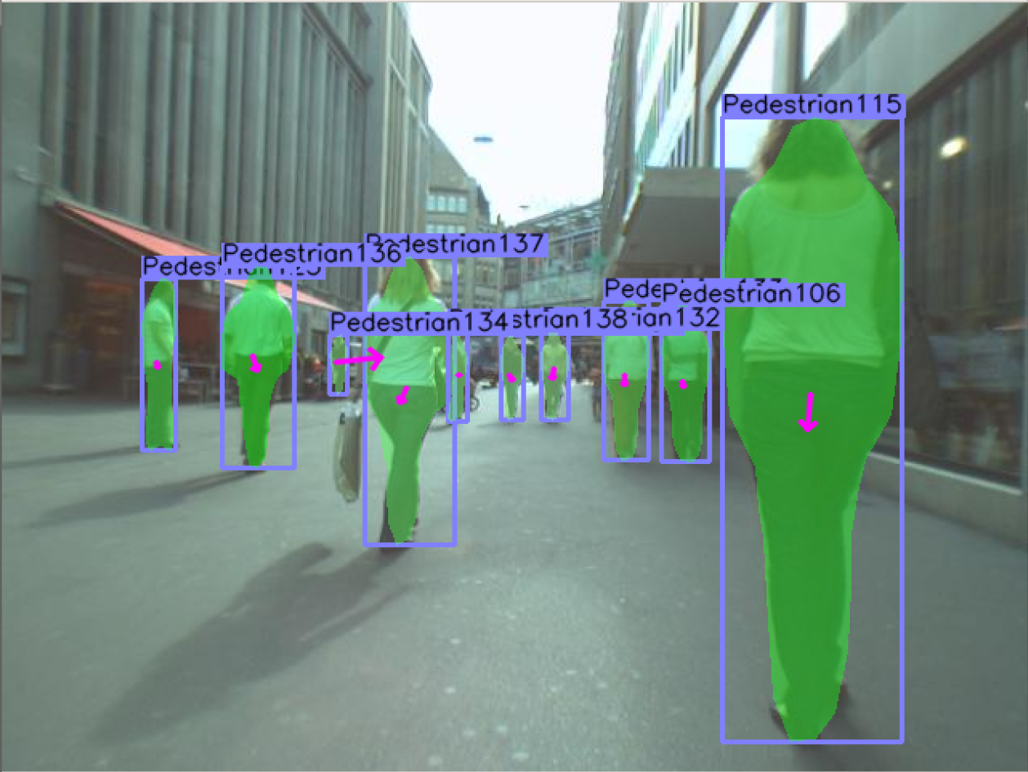}
  \caption{}
  \label{fig:sub2}
\end{subfigure}
\caption{Typical output of PolyTrack. a) A center heatmap to detect objects, b) a bounding polygon for each object, a class label, a track id as well as an offset from the previous frame (pink arrow).}
\label{fig:output}
\end{figure}

For the tracking part of our method, we follow a ``detect then track'' approach and for the segmentation part, we follow a ``detect then segment'' approach. We detect objects by finding their center keypoint, thus we train a network to produce center keypoint heatmaps. Simultaneously, another network head produces a dense polygon map, which contains a bounding polygon at each location on the map. The polygon head is trained such that the locations of object centers on the polygon map contain the polygons corresponding to the same respective objects. Another network head is trained to produce a tracking offset between corresponding objects of two consecutive frames. Using the proximity of the centers of the objects, we associate them into tracklets over whole sequences. A Kalman filter is also applied to reduce the number of ID switches. A typical output of our method is shown in Figure~\ref{fig:output}. Our method, PolyTrack, a fast multi-object tracking and segmentation method was evaluated on the KITTIMOTS~\citep{Voigtlaender19CVPR_MOTS} and MOTS~\citep{Voigtlaender19CVPR_MOTS} datasets. Results show that our proposed method is promising. 

The contribution of our paper is the presentation of a novel MOTS method called PolyTrack, that is based on polygonal masks. 
PolyTrack can be thought of as an improved traditional multi-object tracker providing something better than bounding boxes, that is bounding polygons, at almost no additional cost. It was not designed to provide overly fine segmentation masks, but rather coarse segmentations that are sufficient in most cases (see Figure \ref{fig:sub2}), for instance to remove the background in an image, or provide better re-identification features. PolyTrack obtains good results on the tested benchmarks, particularly for more rigid objects, but not SOTA results. This is expected since PolyTrack does not produce segmentation masks, nor bounding boxes. Therefore, it is not possible to get impressive results with the established MOTS metrics that require 50\% overlap with the ground-truth.

\section{Related Work}
\subsection{Multiple Object Tracking}
 The majority of high performing multiple object trackers follows a tracking by detection paradigm. First, using a pre-trained object detector~\citep{felzenszwalb2009object,ren2017accurate,ren2015faster,yang2016exploit}, they produce bounding boxes for the current video frame. In a second stage, they perform data association to get the tracks by matching the current frame bounding boxes with the previous frames bounding boxes. 

For example, SORT~\citep{bewley2016simple} uses common tracking techniques like the Kalman Filter and the Hungarian algorithm combined with a high quality object detector, Faster R-CNN~\citep{ren2015faster}, to produce good online results. Only bounding box overlap is used. In a follow-up work, Deep SORT~\citep{wojke2017simple}, the appearance is integrated, making the method less susceptible to identity switches. Learning by tracking~\citep{leal2016learning} implements a new two-stage approach to data association while tracking pedestrians. In the first stage, a Siamese convolutional neural network (CNN) is trained to produce descriptor encodings of the input images. Second, contextual information from the position and size of the input image patches is integrated with the CNN by gradient boosting to produce the matching probability. The work of~\citet{schulter2017deep} show that it is possible to learn features for data association in a network-flow via backpropagation. The authors express the optimum of a smoothed network flow problem as a differentiable function. They outperform the methods based on handcrafted features. The authors of ~\citep{tang2017multiple} proposed a method that links person detections over time by solving a minimum cost lifted multicut problem. To group tracks over longer periods of time, they implemented a new deep architecture for re-identification. The work of~\citet{sharma2018beyond} leverages traditional MOT techniques by incorporating 3D information into the cost for data association. \citet{ooi2018multiple} improve data association by using label information provided by the detector. The work of~\citep{xu2019spatial} proposes a unified framework to obtain similarity measures between image patches that can encode different cues and work across both spatial and temporal domains. Some methods, like CenterTrack~\citep{zhou2020tracking} and Tracktor \citep{bergmann2019tracking}, frame multi-object tracking as a regression problem (tracking by regression), using simple strategies for instance association but relying on strong object detectors to get good performance. The detectors are designed to also regress the object displacement. 

\subsection{Multiple Object Tracking and Segmentation}
Multiple object Tracking and Segmentation (MOTS) is a fairly novel compound task that combines instance segmentation and multiple object tracking. Track R-CNN~\citep{Voigtlaender19CVPR_MOTS} is a baseline method proposed for the MOTS dataset. It extends Mask R-CNN~\citep{he2017mask} with 3D convolutions to incorporate the temporal aspect and a network head that performs data association. ReMOTS~\citep{yang2020remots} proposes a self-supervised refinement method by using short-term tracklets as pseudo labels to train for longer ones. GMPHD\_MAF~\citep{song2020online} takes instance segmentation results as input and performs data association based on the Gaussian mixture probability hypothesis density filter for position and motion, and kernelized correlation filter for appearance. UniTrack~\citep{wang2021different} proposes a unified framework for solving five different tasks. It consists of an appearance model that is task agnostic, and several heads which are used for solving each task and do not require training. The appearance model can be either supervised or self-supervised. SORTS~\citep{ahrnbom2021real} presents the first real-time MOTS method. It is based on SORT and uses instance masks produced by Mask R-CNN. Additionally, they present an alternative method called SORTS-RReID which uses a re-identification method to better handle occlusions. 

\section{Proposed Method}

Our method is based on the CenterTrack and CenterPoly models (these two being based on the CenterNet model). Our aim is to design a tracking by regression method to track objects delimited by polygons. The polygons being approximations of masks, we effectively and efficiently transform a bounding box tracking method into a mask tracking method, that is a MOTS method. 

CenterNet is a multi-class object detector that predicts the position of the center of objects in an image using a heatmap and then obtains the coordinates of the bounding boxes of each object by regression. It is an object representation model that offers a lot of potential. Therefore, it is possible to adapt this base model for several purposes. It can be adapted for tracking (CenterTrack) and to regress polygons to get mask approximations (CenterPoly). Our proposed method, PolyTrack, draws ideas from both of these derived methods to design a complete MOTS solution.

\subsection{General architecture of PolyTrack}
\label{genarch}

\begin{figure*}[ht]
\begin{center}
\includegraphics[width=1\linewidth]{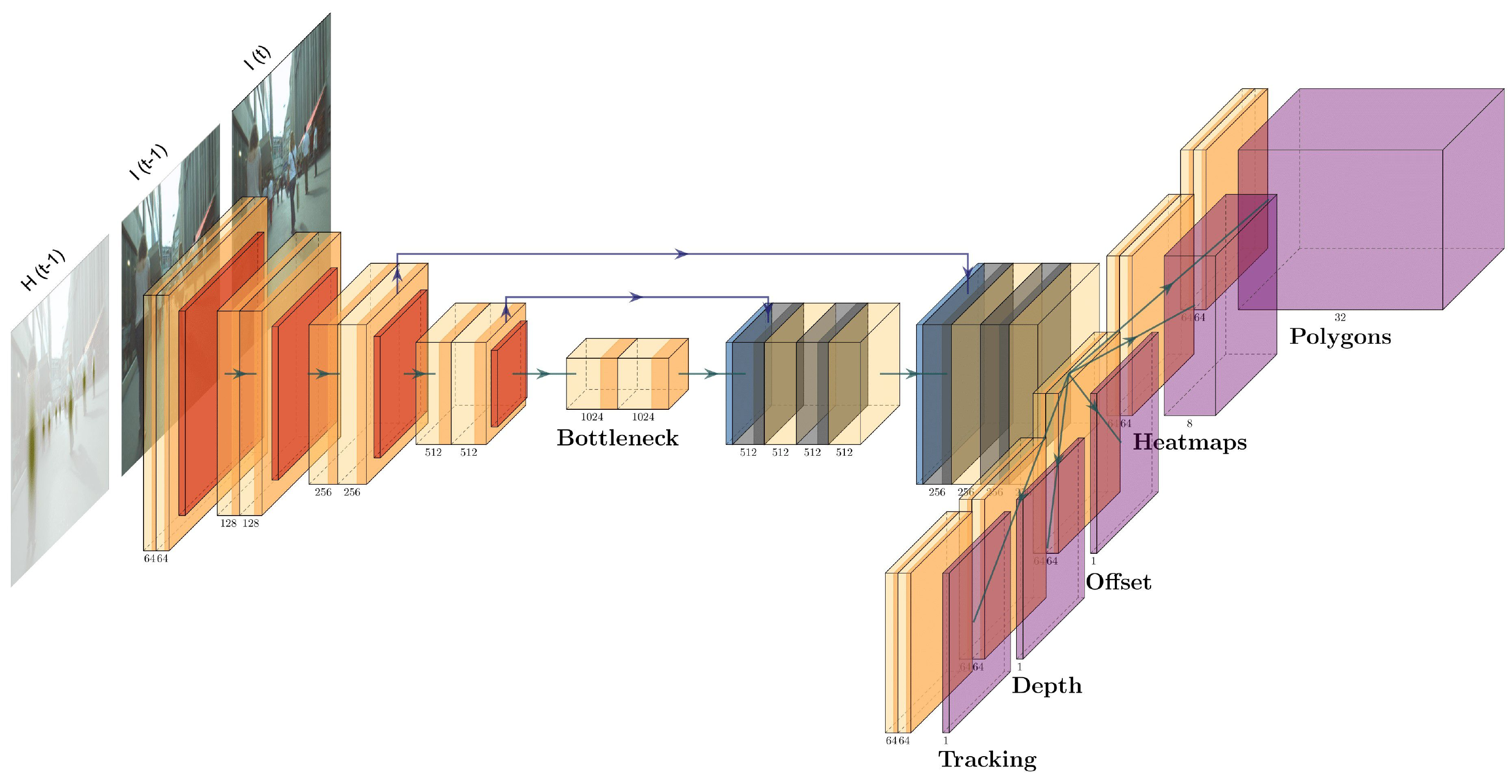}
\end{center}
\caption{An overview of the PolyTrack architecture. The network takes as input the image at time $t$, $I(t)$, the image at time $t-1$, $I(t-1)$, as well as the heatmap at time $t-1$, $H(t-1)$. Features are produced by the backbone and then used by five different network heads. The center heatmap head is used for detecting and classifying objects, the polygon head is used for the segmentation part, the depth head is used to produce a relative depth between objects, the tracking head is used to produce an offset between frames at time $t-1$ and time $t$ and finally the offset head is used for correctly upsampling images. }
\label{polytrack}
\end{figure*}

The network architecture of PolyTrack is presented in Figure~\ref{polytrack}. Similarly to CenterPoly, it has a polygon regression head and a depth head to order the polygons by depth. Also similarly to CenterTrack, it has a tracking head to obtain the displacement of objects between two frames.

There are three inputs to the network: the current image $I(t)$, the previous image $I(t-1)$ and a heatmap $H(t-1)$ representing the positions of the tracklets detected in the previous image $I(t-1)$. These three inputs each pass through a small convolutional network consisting of a convolution layer and a residual layer. These convolutions downsample the input resolution of the inputs by 4 and set the number of channels of the backbone input tensor to 256. The outputs of these three small networks are then summed before being passed through the network backbone. The backbone we used to produce our results is a 2-stacks hourglass. The features generated by the backbone are then used by the regression heads to produce outputs. The regression heads, except the one for polygons, are built on the same model, a $3 \times 3$ convolution followed by a $1\times 1$ convolution that generate one or more output values at each pixel of the downsampled feature map. The polygon regression head greatly benefits from multiple layers and a greater depth. The polygon regression head is therefore composed of three successive blocks of $3\times 3$ convolution, $1\times 1$ convolution and a ReLU activation before generating the output feature map (see code for details). Our architecture is detailed in the following. 

\subsection{Detecting objects as polygons}

To obtain mask approximation of the objects to track, we propose an approach similar to CenterPoly. First, the CenterNet bounding box generation head is replaced by a polygon generation head. Instead of simply generating a bounding box around an object, we generate a bounding polygon that allows for a much finer segmentation. To learn to generate the polygons, the heatmaps provided as ground truth for training are generated by producing elliptical Gaussians whose length and width depend on the object dimensions. This improves detection accuracy compared to circular Gaussians, as shown in~\citep{perreault2021centerpoly}. An object center is no longer defined as the center of its bounding box, but as its center of gravity computed as the weighted mean of the points forming the bounding polygon. 

Finally, similarly to CenterPoly, we use a pseudo-depth head that allows to learn the relative depth of objects in the frame to determine for two overlapping masks which one is in front of the other.

\subsection{Tracking the object polygons}

To track the object polygons, we use a similar approach to CenterTrack. Instead of having a single image as the input of our network, we use the current image, $I(t)$, the previous image $I(t-1)$, and a heatmap $H(t-1)$ generated from the positions of the tracks present in the previous image. We also added a tracking regression head. This tracking head, a bit like an optical flow, predicts for each object detected in the current image their position in the previous image. This information is then used to establish associations between instances of the two frames using a greedy algorithm: an object will be assigned an existing ID if its previous position predicted by the tracking head is similar to the position of the center of an object detected in the previous frame. 

In the CenterTrack method, the association of instances between the frames of a video stream is based solely on the object centers detected by the neural network and the displacement predicted by the tracking head. For each detected object, the displacement vector produced by the tracking head is added to the position of the object center. These new positions are then compared with the positions of the centers of the previously detected objects, and a greedy algorithm is used to associate the centers closest to each other between two frames to propagate the identity of already existing tracks. If a detection in the current frame is not close to any existing track, a new track is created with a new ID. If an existing track is not close to any detection of the current frame, it is frozen and a counter, reflecting for how long a track was not matched, is incremented until either a detection of a next frame is associated with it or its value exceeds a limit (established at 32 frames in our case). If the counter value exceeds the set limit, the track is terminated, as we consider that the instance left the video stream.

This solution has the advantage of being quite simple and efficient. However, when a track is frozen, its position remains as the last detected position. In practice, this has the consequence that the tracklet of an instance $A$ that has been frozen for a few frames because of an occlusion or an overlapping of two instances has a high chance of being ``captured'' by another instance $B$ passing nearby, causing an ID switch. If no instance $B$ passes nearby, but instance $A$ reappears too far from its last known position, another ID switch may happen, the new position being too far from the old one to be associated with it. Finally, if instance $A$ exits the video frame, its last known position will be on the boundary of the frame and its track will remain frozen there until the maximum counter limit is reached. However, if an instance $B$ enters the frame from the same area on the boundary, it will most likely be associated with the last known position of instance $A$, causing an ID switch once again.

To address these three problems and inspired by \citep{chen2018tracking}, we bring a partial solution: the use of Kalman Filters, more precisely Unscented Kalman Filters (UKF). A Kalman filter works by estimating, through observations, the parameters of a system in order to be able to remove noise from observations and to predict a future state of the system. In our case, the system under study corresponds to objects moving through a video stream. Newton's kinematics equations are therefore used to get an approximation of an object state (position, velocity and acceleration) at a time $t$. We use an UKF because it is much more efficient than a classical Kalman filter when applied to nonlinear systems and because it has already been proven efficient in the field of object tracking \citep{chen2018tracking}. An UKF is applied to each track at its birth. As long as a track is active and detected in the video stream, each new position of the instance is used to update the UKF, which allows it to progressively build the kinematic state of the object (position, velocity and acceleration). When a track is frozen, the UKF is used to predict the position of the object in each frame until the track is abandoned, or a new association is made with one of the instance detections in a subsequent frame. When an object disappears from a frame in the video stream, the tracker assumes that it will continue to move on the same trajectory it was going before disappearing.

This use of the UKF thus allows us to partially counter the problems presented above and to significantly reduce the number of identity switches happening at inference. When an object leaves the frame of the video stream, the UKF will make it continue on its way and therefore its position for the tracker will be outside the dimensions of the frame, so a new instance entering the frame afterwards is less likely be associated with it. In the same way, a track that has been frozen because of an occlusion will have a better chance of being correctly associated with its original instance if, when it reappears, its position is approximately predicted to follow the true position of the instance.

\subsection{Training loss}

The loss functions used for each of the outputs of the regression heads are a Focal Loss~\citep{lin2017focal} for the center point heatmap generation head and a simple $L1$ Loss for each of the other regression heads. The global loss is a linear combination of the different losses of each head. It is given by

\begin{equation}
    L_{tot}=w_{hm}\ast L_{hm}+w_{poly}\ast L_{poly}+w_{depth}\ast L_{depth}+w_{track}\ast L_{track}+w_{off}\ast L_{off},
    \label{eq_loss}
\end{equation}

where $L_{hm}$ is the focal loss as defined by CenterNet, $L_{poly}$, $L_{depth}$, $L_{track}$ and $L_{off}$ are the $L1$ losses as defined by CenterPoly and CenterTrack. At training, we used weights of 1 for $ w_{HM}, w_{poly}, w_{track}$ and $w_{off}$, and of 0.1 for $w_{depth}$. 

We use the same vertex selection policy for ground truth polygon generation as described in detail in \citep{perreault2021centerpoly}. To summarize it, it consists in drawing lines at regular intervals around the bounding box of the instance towards its center and to keep the first point of each line that intersects with the instance segmentation mask. The polygons learned by the model correspond to the coordinates of these vertices as offsets from the object center. The object centers are moreover not defined as the center of the bounding box, but as an average of the coordinates forming the bounding polygon, which makes it easier to learn the vertices as offsets. The ground truth regarding the tracking head is quite simple, for each object center $(x_{t}, y_{t})$ present in an image $I(t)$, if the same object is in the image $I(t-1)$ at $(x_{t-1}, y_{t-1})$, the ground truth for the object displacement corresponds to 

\begin{equation}
    (x_{off},y_{off}\ )=(x_{t-1},\ y_{t-1}\ )-(x_{t},y_{t}).
    \label{eq_off}
\end{equation}

The loss of all outputted downsampled feature maps (except for the center point heatmap) is only evaluated at the centers of object instances present in the current image.

\section{Experiments}

We tested PolyTrack on two datasets, MOTS and KITTIMOTS. We also performed an ablation studies to assess the contribution of the various parts of our method.

\subsection{Datasets used in the experiments}

MOTS~\citep{Voigtlaender19CVPR_MOTS} is a small dataset whose training set consists of only four videos and one class, pedestrians. The challenge of using polygons is that when generating the ground truth, a lot of information is already lost and the resulting masks are therefore already not perfect. For a dataset like MOTS whose results are essentially evaluated by metrics based on IoU, MOTSA and sMOSTA, it is simply impossible to obtain a perfect score when evaluating with these metrics. In Table~\ref{table_mots}, the first row shows the maximum scores obtained by evaluating the ground truth approximated by polygons against the ground truth provided with the dataset, i.e.\ our upper bound. We used MOTS as the basis for running our experiments and testing the incremental improvements made to the model. We separated the dataset into a training set consisting of the videos MOTS20-02, MOTS20-05, and MOTS20-11 and a validation set consisting of video MOTS20-09. The results in Table~\ref{table_mots} are from experiments conducted on this validation set. Results on the test set are in Table~\ref{results-mots}.

KITTIMOTS~\citep{Voigtlaender19CVPR_MOTS} is a bigger dataset evaluated on cars and pedestrians that contains 21 training and 29 test sequences in urban environment. Methods are evaluated and ranked by HOTA~\citep{Luiten2020IJCV}. Our results compared with published and peer-reviewed state-of-the-art methods are shown in Table~\ref{results-kitti} for cars and pedestrians. We also show a study made on the usage of UKF on a custom train/evaluation split on KITTIMOTS in Table~\ref{results-ukf}.

\subsection{Details about the training}

We realized during our experiments that it was almost impossible to start from randomly initialized weights for a trained model to be efficient. Indeed, when training a model from scratch, it never really seemed to converge for a simple reason: the network gives too much importance to the input heatmap and too little to the two other inputs, $I(t)$ and $I(t-1)$. This means that the network is then unable to detect objects without already existing tracklets. This is illustrated by the second row of Table \ref{table_mots}. Track births are therefore near impossible since at the beginning of a video sequence the heatmap passed in parameter is empty, the network must then only rely on the images $I(t)$ and $I(t-1)$ to detect the instances. The solution to the problem is to use a pre-trained network. Therfore, during training, we fine-tuned our models from weights of a CenterNet backbone (which only rely on images to perform detection) trained on the COCO~\citep{DBLP:journals/corr/LinMBHPRDZ14} dataset. 

Regarding the model training parameters, we have kept the optimal parameters determined by the CenterTrack authors. Polygons have 32 vertices. We applied standard data augmentation: image flipping, translations, rotations and color shifts. We also introduced perturbations to the input heatmap at training to make it more resilient towards errors at inference: the generated instance centers are slightly shifted from their position in the ground truth, there is a 40\% chance that an object center is omitted to simulate a detection error and there is a 10\% chance that a false positive is added. The input image $I(t-1$) is also randomly chosen from the three frames preceding the current frame $I(t)$ to make sure the model does not overfit on a single framerate. 

\subsection{Results and discussion}

\begin{table}[]
\scriptsize
\caption{Results of various tracker configuration on our validation set from the MOTS~\citep{Voigtlaender19CVPR_MOTS} dataset. We tested two backbones: DLA34 and HG (2-stacks hourglass), W: using pre-trained weights for the backbone,  Deep: adding 2 additional layers to the polygons regression head, Maximum: Perfect detection and tracking with polygons.}
\vspace{5pt}
\centering
\begin{tabular}{l|l|l|l|l|l|l|l|l|l}
              & sMOTSA & MOTSA & MOTSP & TP   & FP  & FN   & Rcll  & Prcn  & IDSWR \\
\hline
Maximum       & 71.97  & 89.82 & 81.19 & 4531 & 243 & 243  & 94.91 & 94.91 & 0     \\
DLA34         & 12.11  & 25.58 & 69.13 & 2083 & 829 & 2691 & 43.63 & 71.53 & 75.6  \\
DLA34 + W     & 21.11  & 34.27 & 68.8  & 2014 & 341 & 2760 & 42.19 & 85.52 & 87.7  \\
HG + W        & 27.74  & 47.55 & 68.85 & 3036 & 720 & 1738 & 63.59 & 80.83 & 72.3  \\
HG + W + Deep & 32.88  & 51.59 & 69.01 & 2882 & 383 & 1892 & 60.37 & 88.27 & 59.6 \\
\end{tabular}
\label{table_mots}
\end{table}

\begin{table}
\scriptsize
\caption{Results on the KITTIMOTS~\citep{Voigtlaender19CVPR_MOTS} dataset for the Car and pedestrian categories. \textbf{Bold} means best result. }
\centering
\textbf{Cars}\\
\vspace{2pt}
\begin{tabular}{l|c|c|c|c|c|c|c|c|c}
\hline
{\bf Method} & {\bf HOTA} & {\bf DetA} & {\bf AssA} & {\bf DetRe} & {\bf DetPr} & {\bf AssRe} & {\bf AssPr} & {\bf LocA} & {\bf sMOTSA}\\ \hline
ViP-DeepLab~\citep{vip_deeplab} & \textbf{76.38\%} & \textbf{82.70\%} & 70.93\% & \textbf{88.70\%} & 88.77\% & 75.86\% & 86.00\% & \textbf{90.75\%} & \textbf{81.03\%}\\
EagerMOT~\citep{Kim21ICRA} & 74.66\% & 76.11\% & \textbf{73.75\%} & 79.59\% & 90.24\% & \textbf{76.27\%} & \textbf{92.70\%} & 90.46\% & 74.53\%\\
MOTSFusion~\citep{luiten2019MOTSFusion} & 73.63\% & 75.44\% & 72.39\% & 78.32\% & \textbf{90.78\%} & 75.53\% & 89.97\% & 90.29\% & 74.98\%\\
ReMOTS~\citep{yang2020remots} & 71.61\% & 78.32\% & 65.98\% & 83.51\% & 87.42\% & 68.03\% & 92.61\% & 89.33\% & 75.92\%\\
PointTrack~\citep{xu2020Segment} & 61,95\% & 79,38\% & 48,83\% & 85,77\% & 85.66\% & 79.07\% & 56.35\% & 88.52\% & 78.50\%\\
PolyTrack (ours) & 57.61\% & 62.47\% & 53.97\% & 67.83\% & 79.11\% & 57.13\% & 81.92\% & 83.70\% & 57.49\%\\
TrackR-CNN~\citep{Voigtlaender19CVPR_MOTS} & 56.63\% & 69.90\% & 46.53\% & 74.63\% & 84.18\% & 63.13\% & 62.33\% & 86.60\% & 66.97\%\\
GMPHD\_SAF~\citep{gmphdsaf} & 55.14\% & 77.01\% & 39.76\% & 81.57\% & 87.29\% & 69.22\% & 49.42\% & 88.72\% & 75,39\%
\end{tabular}
\centering
\vspace{2pt}
\textbf{Pedestrians}\\

\begin{tabular}{l|c|c|c|c|c|c|c|c|c}
\hline
{\bf Method} & {\bf HOTA} & {\bf DetA} & {\bf AssA} & {\bf DetRe} & {\bf DetPr} & {\bf AssRe} & {\bf AssPr} & {\bf LocA} & {\bf sMOTSA}\\ \hline
ViP-DeepLab~\citep{vip_deeplab} & \textbf{64.31\%} & \textbf{70.69\%} & \textbf{59.48\%} & \textbf{75.71\%} & 81.77\% & \textbf{67.52\%} & 74.92\% & \textbf{84.40\%} & \textbf{68.76\%}\\
ReMOTS~\citep{yang2020remots} & 58.81\% & 67.96\% & 52.38\% & 71.86\% & \textbf{82.22\%} & 54.40\% & \textbf{88.23\%} & 84.18\% & 65.97\%\\
EagerMOT~\citep{Kim21ICRA} & 57.65\% & 60.30\% & 56.19\% & 63.45\% & 81.58\% & 60.19\% & 83.35\% & 83.65\% & 58.08\%\\
PointTrack~\citep{xu2020Segment} & 54.44\% & 62.29\% & 48.08\% & 65.49\% & 81.17\% & 64.97\% & 58.66\% & 83.28\% & 61.47\%\\
MOTSFusion~\citep{luiten2019MOTSFusion} & 54.04\% & 60.83\% & 49.45\% & 64.13\% & 81.47\% & 56.68\% & 70.44\% & 83.71\% & 58.75\%\\
GMPHD\_SAF~\citep{gmphdsaf} & 49.33\% & 65.45\% & 38.32\% & 69.62\% & 80.98\% & 57.88\% & 49.77\% & 83.82\% & 62.87\%\\
PolyTrack (ours)& 44.89\% & 48.63\% & 41.82\% & 52.78\% & 66.32\% & 46.70\% & 63.05\% & 74.40\% & 39.75\%\\
TrackR-CNN~\citep{Voigtlaender19CVPR_MOTS} & 41.93\% & 53.75\% & 33.84\% & 57.85\% & 72.51\% & 45.30\% & 50.74\% & 78.03\% & 47.31\%
\end{tabular}

\label{results-kitti}
\end{table}

\begin{table}
\scriptsize
\caption{Results on the MOTS~\citep{Voigtlaender19CVPR_MOTS} dataset. \textbf{Bold} means best result. }
\vspace{5pt}
\begin{tabular}{l|c|c|c|c|c|c|c|c|c}
{\bf Method} & {\bf sMOTSA} & {\bf IDF1} & {\bf MOTSA} & {\bf MOTSP} & {\bf MODSA} & {\bf Rcll} & {\bf Prcn} & {\bf IDSw} & {\bf Hz}\\ \hline
ReMOTS~\citep{yang2020remots} & \textbf{70,4}\% & \textbf{75.0}\% & \textbf{84.4}\% & 84.0\% & \textbf{85.1\%} & 87.6\% & \textbf{97.2\%} & \textbf{231} & 0.3\\
GMPHD\_MAF~\citep{song2021online} & 69.4\% & 66.4\% & 83.3\% & \textbf{84.2}\% & 84.8\% & \textbf{87.7}\% & 96.8\% & 484 & 2.6\\
SORTS~\citep{5f6bf791d5cd448e8afe5e578a356583} & 55.0\% & 57.3\% & 68.3\% & 81.9\% & 70.0\% & 73.4\% & 95.7\% & 552 & \textbf{54.5}\\
TrackR-CNN~\citep{Voigtlaender19CVPR_MOTS} & 40.6\% & 42.4\% & 55.2\% & 76.1\% & 56.9\% & 60.8\% & 94.0\% & 567 & 2.0\\
PolyTrack (ours)& 30.7\% & 50.8\% & 48.5\% & 70.1\% & 49.5\% & 59.6\% & 85.5\% & 340 & 9.1\\
\end{tabular}

\label{results-mots}
\end{table}

\begin{table}
\scriptsize
\caption{Results of the UKF study on the KITTIMOTS~\citep{Voigtlaender19CVPR_MOTS} train/val split. }
\vspace{5pt}
\begin{tabular}{l|c|c|c|c|c|c|c|c|c}
{\bf} & {\bf HOTA} & {\bf DetA} & {\bf AssA} & {\bf DetRe} & {\bf DetPr} & {\bf AssRe} & {\bf AssPr} & {\bf LocA} & {\bf sMOTSA}\\ \hline
Cars w/ UKF & 64,25\% & 62,40\% & 66,84\% & 68,22\% & 77,82\% & 70,68\% & 82,15\% & 83,13\% & 57,42\%\\
Cars & 59,80\% & 57,12\% & 63,31\% & 62,05\% & 76,87\% & 67,40\% & 79,25\% & 82,26\% & 51,92\%\\
Pedestrians w/ UKF & 36,33\% & 37,41\% & 35,49\% & 42,26\% & 55,24\% & 41,10\% & 53,30\% & 69,89\% & 15,13\%\\
Pedestrians & 32,33\% & 36,13\% & 29,19\% & 41,51\% & 52,73\% & 34,32\% & 46,94\% & 69,79\% & 7,118\%\\

\end{tabular}

\label{results-ukf}
\end{table}

\begin{figure*}[t]
        \centering
        \begin{subfigure}[b]{0.32\textwidth}
            \centering
            \includegraphics[width=\textwidth]{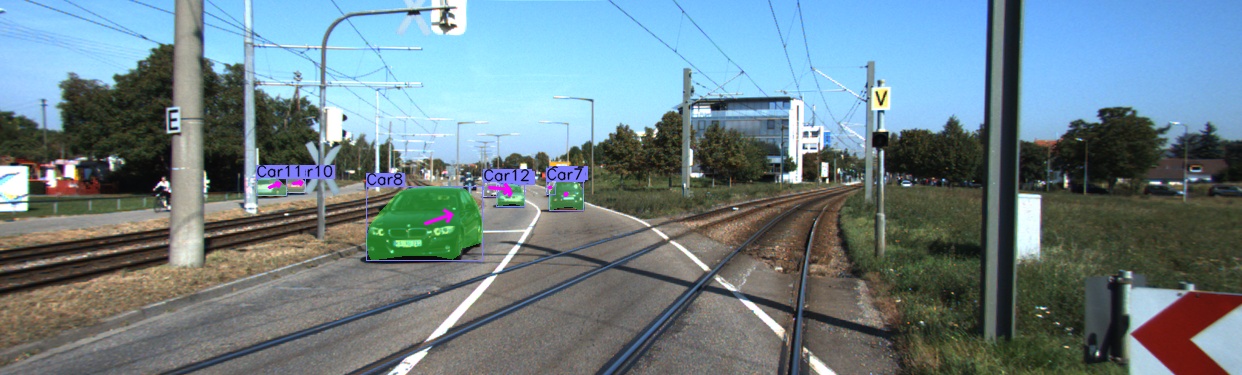}
        \end{subfigure}
        \begin{subfigure}[b]{0.32\textwidth}  
            \centering 
            \includegraphics[width=\textwidth]{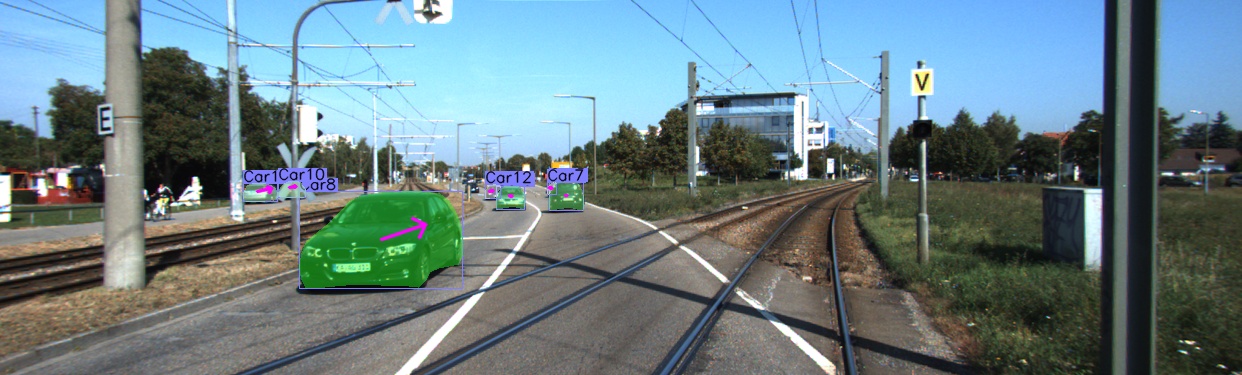}
        \end{subfigure}
        \begin{subfigure}[b]{0.32\textwidth}
            \centering
            \includegraphics[width=\textwidth]{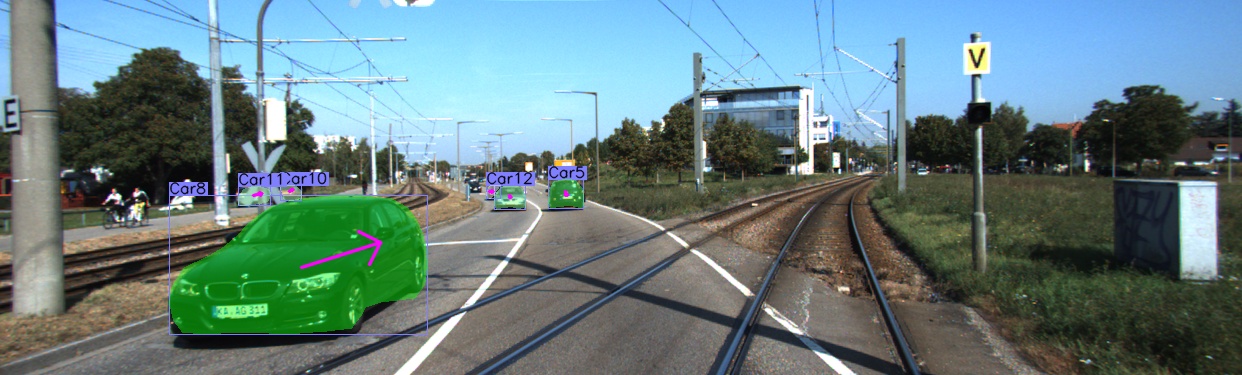}
        \end{subfigure}
        \vskip\baselineskip
        \begin{subfigure}[b]{0.32\textwidth}
            \centering
            \includegraphics[width=\textwidth]{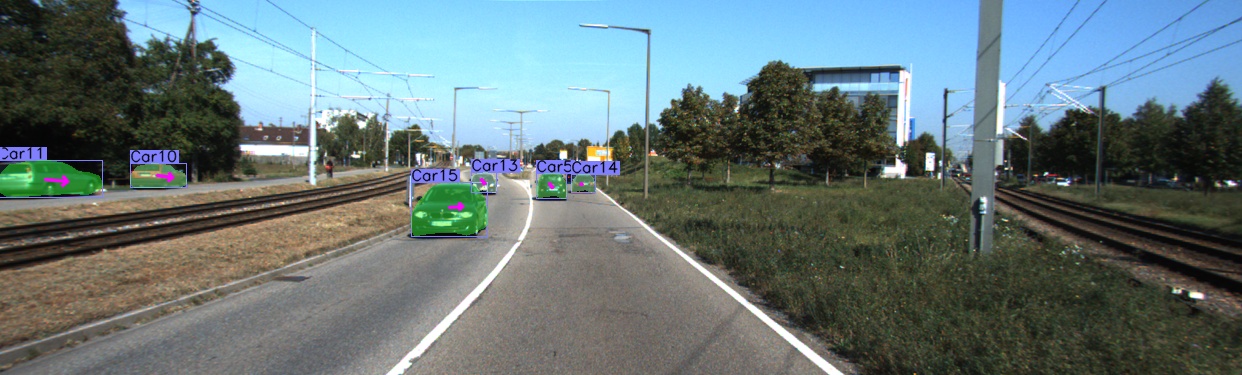}
        \end{subfigure}
        \begin{subfigure}[b]{0.32\textwidth}  
            \centering 
            \includegraphics[width=\textwidth]{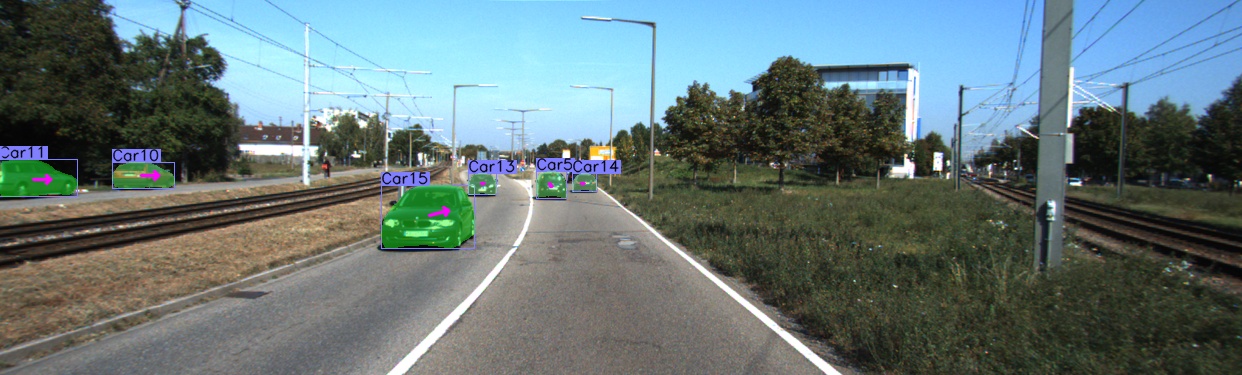}
        \end{subfigure}
        \begin{subfigure}[b]{0.32\textwidth}
            \centering
            \includegraphics[width=\textwidth]{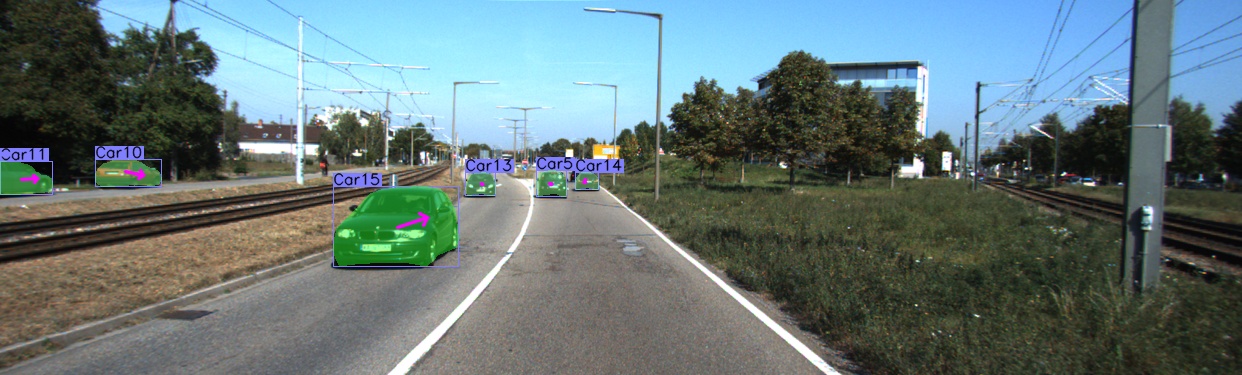}
        \end{subfigure}
        \vskip\baselineskip
        \begin{subfigure}[b]{0.32\textwidth}
            \centering
            \includegraphics[width=\textwidth]{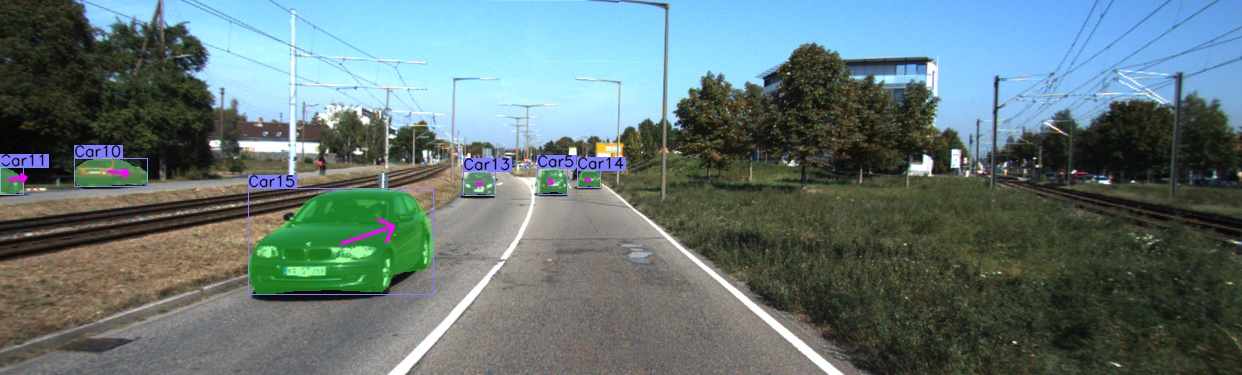}
        \end{subfigure}
        \begin{subfigure}[b]{0.32\textwidth}  
            \centering 
            \includegraphics[width=\textwidth]{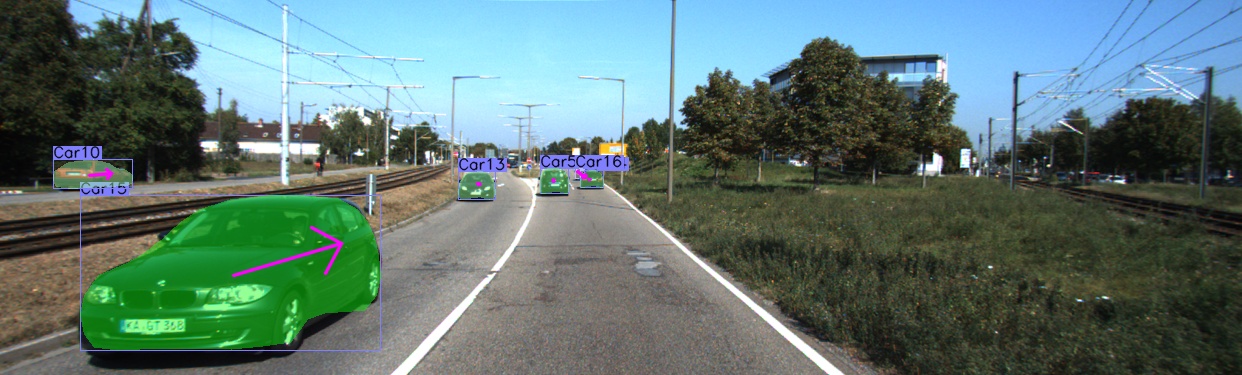}
        \end{subfigure}
        \begin{subfigure}[b]{0.32\textwidth}
            \centering
            \includegraphics[width=\textwidth]{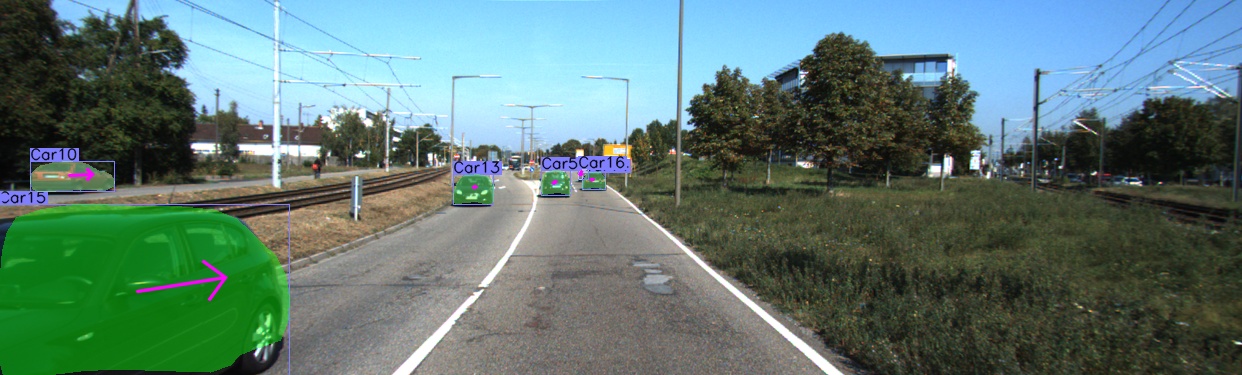}
        \end{subfigure}
        \vskip\baselineskip
        \begin{subfigure}[b]{0.32\textwidth}
            \centering
            \includegraphics[width=\textwidth]{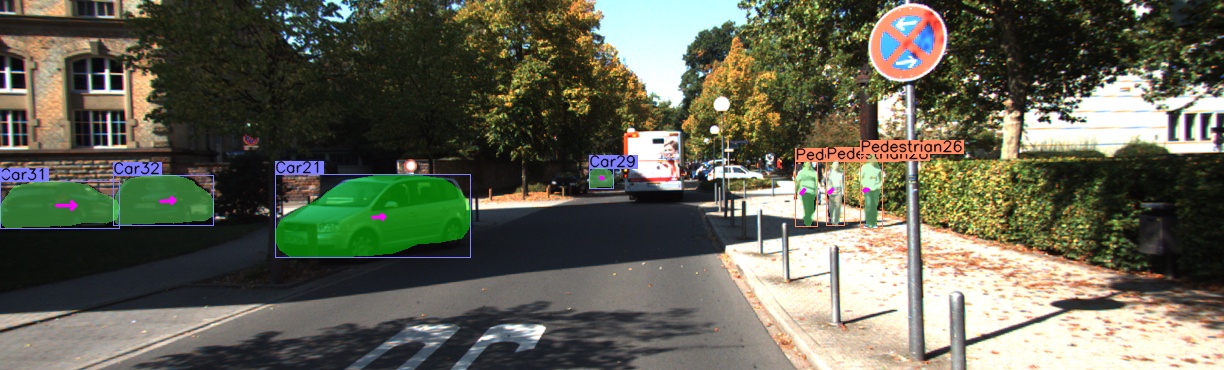}
        \end{subfigure}
        \begin{subfigure}[b]{0.32\textwidth}  
            \centering 
            \includegraphics[width=\textwidth]{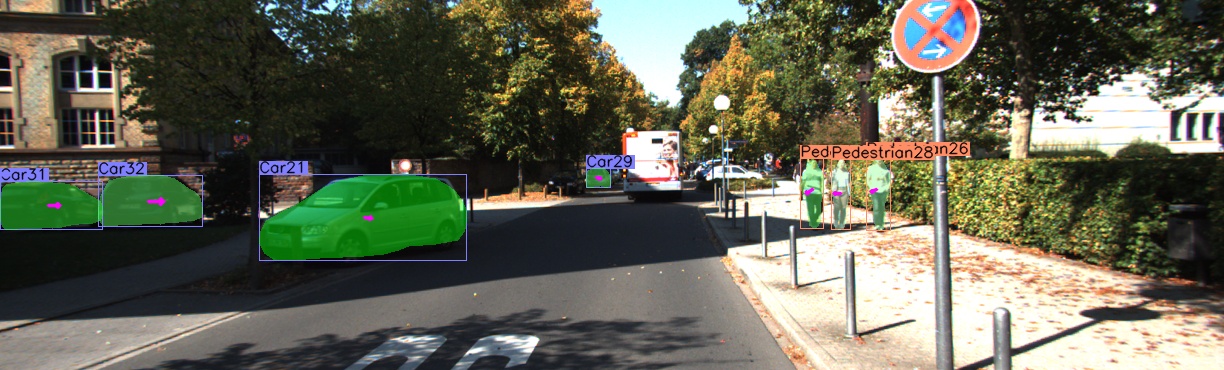}
        \end{subfigure}
        \begin{subfigure}[b]{0.32\textwidth}
            \centering
            \includegraphics[width=\textwidth]{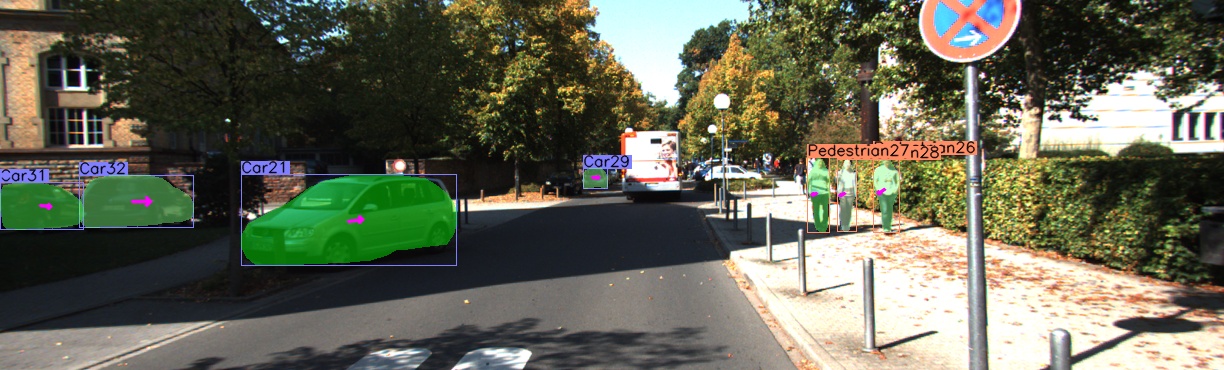}
        \end{subfigure}
        \vskip\baselineskip
        \begin{subfigure}[b]{0.32\textwidth}
            \centering
            \includegraphics[width=\textwidth]{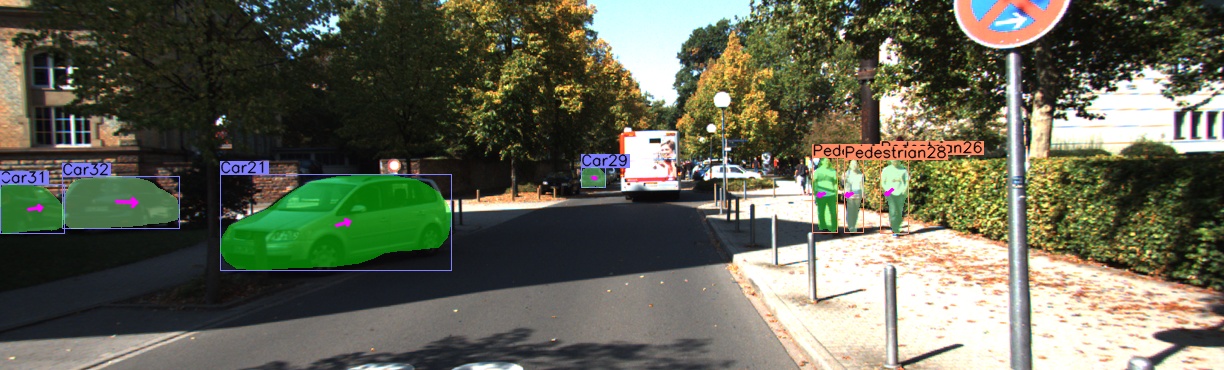}
        \end{subfigure}
        \begin{subfigure}[b]{0.32\textwidth}  
            \centering 
            \includegraphics[width=\textwidth]{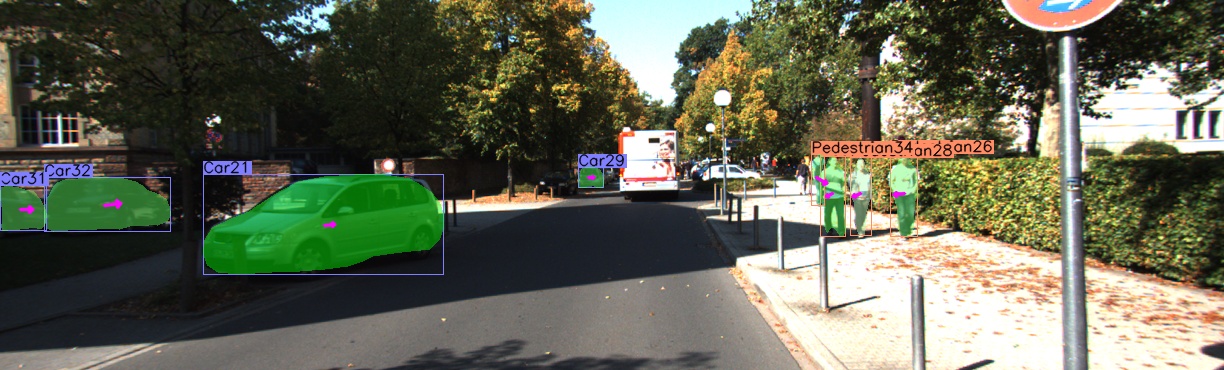}
        \end{subfigure}
        \begin{subfigure}[b]{0.32\textwidth}
            \centering
            \includegraphics[width=\textwidth]{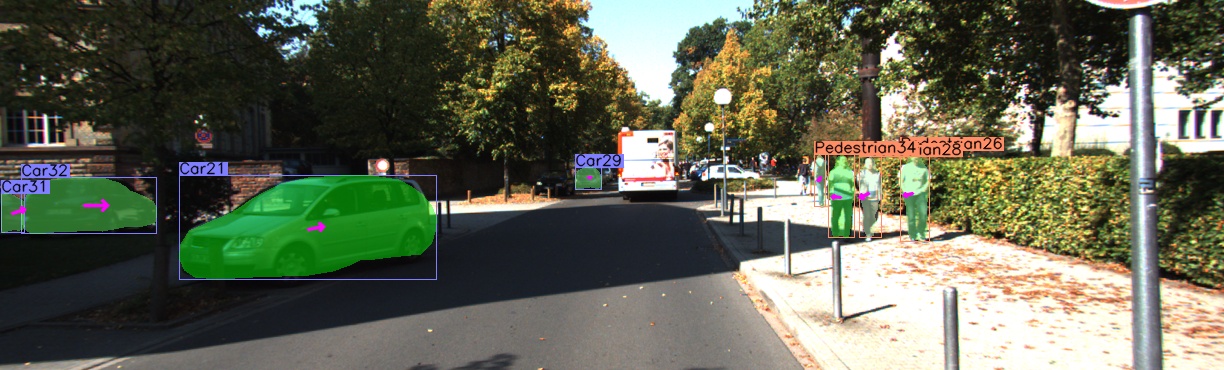}
        \end{subfigure}
        \vskip\baselineskip
        \begin{subfigure}[b]{0.32\textwidth}
            \centering
            \includegraphics[width=\textwidth]{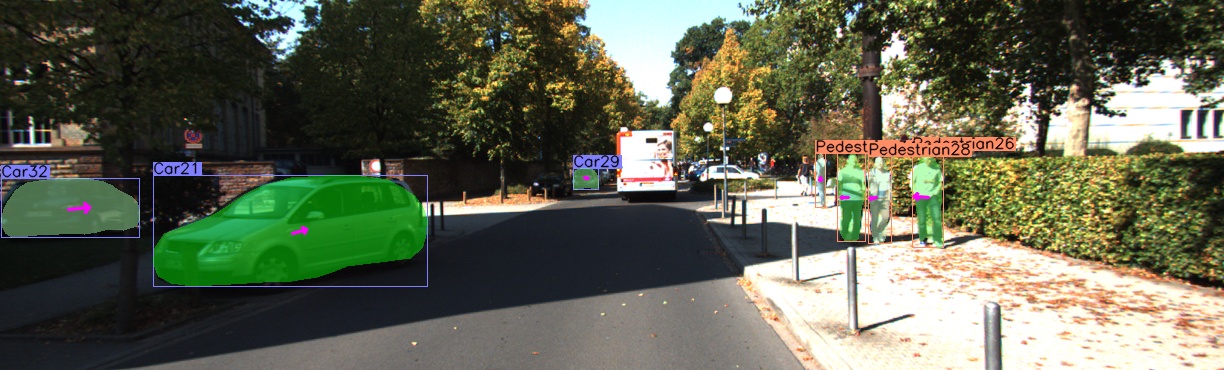}
        \end{subfigure}
        \begin{subfigure}[b]{0.32\textwidth}  
            \centering 
            \includegraphics[width=\textwidth]{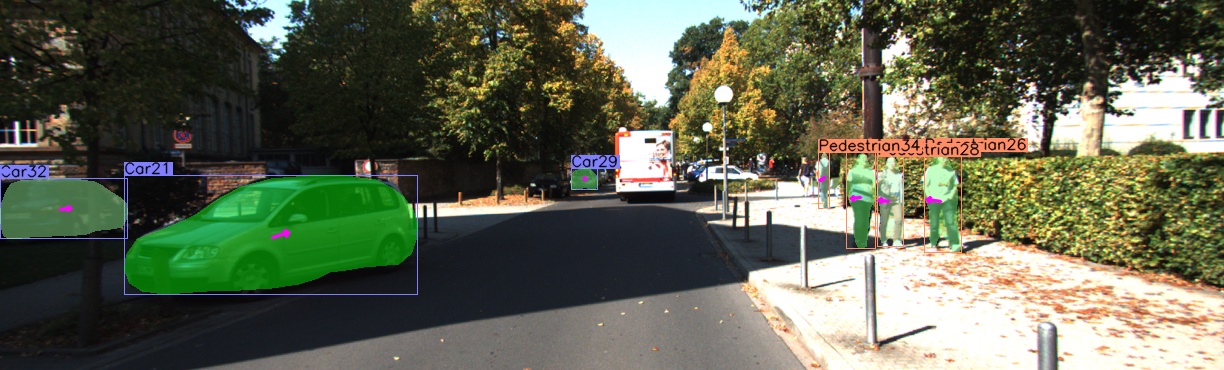}
        \end{subfigure}
        \begin{subfigure}[b]{0.32\textwidth}
            \centering
            \includegraphics[width=\textwidth]{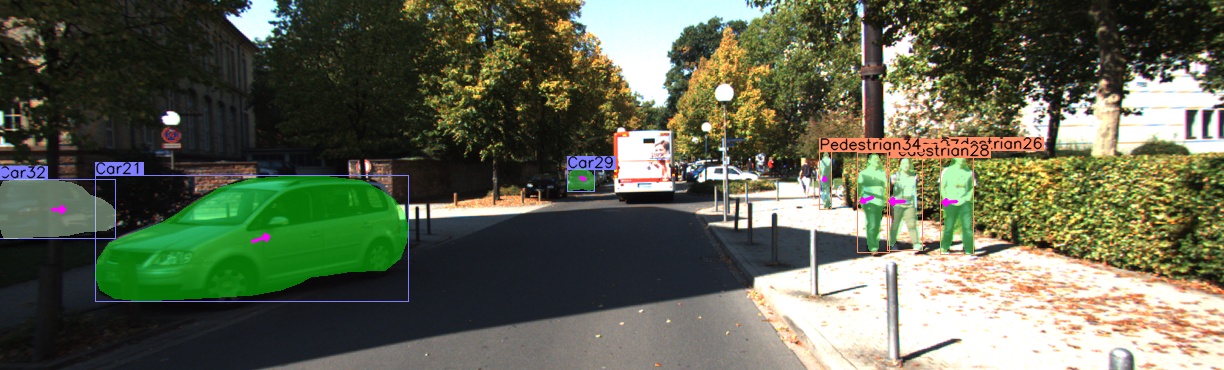}
        \end{subfigure}
        \caption{Qualitative results of PolyTrack on the KITTIMOTS~\citep{Voigtlaender19CVPR_MOTS} dataset. The number represents the instance, the arrow represents the movement between frames and the mask represents the polygon.} 
        \label{qualitative}
    \end{figure*}

Table \ref{table_mots} shows the contribution of various elements of our method. First, it shows how important it is to use pre-trained weights for the backbone network. For DLA-34, we used weights from a CenterPoly model trained on the Cityscapes dataset and for the 2-stacks hourglass we used weights from a CenterNet model trained on the COCO dataset. We can also see that using a 2-stacks hourglass backbone greatly improves the polygon accuracy for segmentation. Finally, we realized that using more layers in the polygon regression head also improved the segmentation accuracy. This is why we used the 2-stacks hourglass backbone with pre-trained weights and a deeper polygons regression head as described in Section \ref{genarch} for generating our final results on MOTS and KITTIMOTS.

Let us now consider Table \ref{results-kitti}. Globally, even though it tracks well and the rough segmentation are good (see figure~\ref{qualitative} for some qualitative results on the KITTIMOTS dataset), PolyTrack suffers from the evaluation metrics when compared to state-of-the-art methods. These metrics are thought of for finer segmentation methods, being quite strict in terms of IOU. This is why our method has the lowest detection recall (DetRe). PolyTrack resides in between bounding box tracking and fine segmentation tracking, providing a bounding polygon instead of a bounding box at almost no additional cost. When compared to other methods, PolyTrack performs better with cars than pedestrians (Table~\ref{results-kitti}). This can be explained by the fact that cars are a lot less deformable than pedestrians, and therefore their representation is easier to learn with polygons. Also, pedestrians are more concave than cars, making them harder to segment with a fixed number of points. For instance, when looking at the detection precision (DetPr), PolyTrack has approximately $6\%$ points less than PointTrack~\citep{xu2020Segment} for the Car category, and approximately $15\%$ points less on the same metric for the Pedestrian category. The results in Table~\ref{results-mots} confirm this trend with passable results on MOTS where only pedestrians are to be tracked. We also believe that MOTS is too small to learn to appropriately extract polygons and that KITTIMOTS reflects better the potential of our method.   

Table~\ref{results-ukf} shows that the UKF improves the tracking significantly for both cars and pedestrians. However, our approach still has its limitations, as the association is still only based on the position of the object center. If two instances cross each other, an identity switch is always possible if the positions of the two instances from one frame to the other correspond exactly at the time of the crossing. The greedy algorithm can then very well swap the two identities at the time of the association, even more so if more than two objects cross at the same time. An improvement of the algorithm based on the UKF could be to include the information of the velocity and acceleration vectors produced by the UKF in the greedy association algorithm. This would allow avoiding potential identity switches between two objects crossing each other (which would therefore have opposite velocity vectors).

\section{Conclusion}
In this work, we presented PolyTrack, a novel multi-object tracking and segmentation method that tracks objects using a detect then track paradigm. PolyTrack detects objects by their center keypoint, segments them using bounding polygons and tracks them with offset regression and an Unscented Kalman filter. PolyTrack shows promising results on MOTS and KITTIMOTS, especially considering the speed / accuracy trade-off.  
%\section{References}
\small
\bibliographystyle{abbrvnat}
\bibliography{bib.bib}

\begin{thebibliography}{32}
\providecommand{\natexlab}[1]{#1}
\providecommand{\url}[1]{\texttt{#1}}
\expandafter\ifx\csname urlstyle\endcsname\relax
  \providecommand{\doi}[1]{doi: #1}\else
  \providecommand{\doi}{doi: \begingroup \urlstyle{rm}\Url}\fi

\bibitem[Ahrnbom et~al.(2021{\natexlab{a}})Ahrnbom, Nilsson, and
  Ard{\"o}]{5f6bf791d5cd448e8afe5e578a356583}
M.~Ahrnbom, M.~Nilsson, and H.~Ard{\"o}.
\newblock Real-time and online segmentation multi-target tracking with track
  revival re-identification.
\newblock In \emph{Proceedings of the 16th International Joint Conference on
  Computer Vision, Imaging and Computer Graphics Theory and Applications},
  volume~5, pages 777--784, 2021{\natexlab{a}}.
\newblock \doi{10.5220/0010190907770784}.

\bibitem[Ahrnbom et~al.(2021{\natexlab{b}})Ahrnbom, Nilsson, and
  Ard{\"o}]{ahrnbom2021real}
M.~Ahrnbom, M.~G. Nilsson, and H.~Ard{\"o}.
\newblock Real-time and online segmentation multi-target tracking with track
  revival re-identification.
\newblock In \emph{VISIGRAPP (5: VISAPP)}, pages 777--784, 2021{\natexlab{b}}.

\bibitem[Bergmann et~al.(2019)Bergmann, Meinhardt, and
  Leal-Taixe]{bergmann2019tracking}
P.~Bergmann, T.~Meinhardt, and L.~Leal-Taixe.
\newblock Tracking without bells and whistles.
\newblock In \emph{Proceedings of the IEEE/CVF International Conference on
  Computer Vision}, pages 941--951, 2019.

\bibitem[Bewley et~al.(2016)Bewley, Ge, Ott, Ramos, and
  Upcroft]{bewley2016simple}
A.~Bewley, Z.~Ge, L.~Ott, F.~Ramos, and B.~Upcroft.
\newblock Simple online and realtime tracking.
\newblock In \emph{2016 IEEE international conference on image processing
  (ICIP)}, pages 3464--3468. IEEE, 2016.

\bibitem[Chen et~al.(2018)Chen, Wang, and Xuan]{chen2018tracking}
X.~Chen, X.~Wang, and J.~Xuan.
\newblock Tracking multiple moving objects using unscented kalman filtering
  techniques, 2018.

\bibitem[Felzenszwalb et~al.(2009)Felzenszwalb, Girshick, McAllester, and
  Ramanan]{felzenszwalb2009object}
P.~F. Felzenszwalb, R.~B. Girshick, D.~McAllester, and D.~Ramanan.
\newblock Object detection with discriminatively trained part-based models.
\newblock \emph{IEEE transactions on pattern analysis and machine
  intelligence}, 32\penalty0 (9):\penalty0 1627--1645, 2009.

\bibitem[He et~al.(2017)He, Gkioxari, Doll{\'a}r, and Girshick]{he2017mask}
K.~He, G.~Gkioxari, P.~Doll{\'a}r, and R.~Girshick.
\newblock Mask r-cnn.
\newblock In \emph{Proceedings of the IEEE international conference on computer
  vision}, pages 2961--2969, 2017.

\bibitem[Kim et~al.(2021)Kim, O\v{s}ep, and Leal-Taix{'e}]{Kim21ICRA}
A.~Kim, A.~O\v{s}ep, and L.~Leal-Taix{'e}.
\newblock Eagermot: 3d multi-object tracking via sensor fusion.
\newblock In \emph{IEEE International Conference on Robotics and Automation
  (ICRA)}, 2021.

\bibitem[Leal-Taix{\'e} et~al.(2016)Leal-Taix{\'e}, Canton-Ferrer, and
  Schindler]{leal2016learning}
L.~Leal-Taix{\'e}, C.~Canton-Ferrer, and K.~Schindler.
\newblock Learning by tracking: Siamese cnn for robust target association.
\newblock In \emph{Proceedings of the IEEE Conference on Computer Vision and
  Pattern Recognition Workshops}, pages 33--40, 2016.

\bibitem[Lin et~al.(2014)Lin, Maire, Belongie, Bourdev, Girshick, Hays, Perona,
  Ramanan, Doll{\'{a}}r, and Zitnick]{DBLP:journals/corr/LinMBHPRDZ14}
T.~Lin, M.~Maire, S.~J. Belongie, L.~D. Bourdev, R.~B. Girshick, J.~Hays,
  P.~Perona, D.~Ramanan, P.~Doll{\'{a}}r, and C.~L. Zitnick.
\newblock Microsoft {COCO:} common objects in context.
\newblock \emph{CoRR}, abs/1405.0312, 2014.
\newblock URL \url{http://arxiv.org/abs/1405.0312}.

\bibitem[Lin et~al.(2017)Lin, Goyal, Girshick, He, and
  Doll{\'a}r]{lin2017focal}
T.-Y. Lin, P.~Goyal, R.~Girshick, K.~He, and P.~Doll{\'a}r.
\newblock Focal loss for dense object detection.
\newblock In \emph{Proceedings of the IEEE international conference on computer
  vision}, pages 2980--2988, 2017.

\bibitem[Luiten et~al.(2020{\natexlab{a}})Luiten, Fischer, and
  Leibe]{luiten2019MOTSFusion}
J.~Luiten, T.~Fischer, and B.~Leibe.
\newblock Track to reconstruct and reconstruct to track.
\newblock \emph{IEEE Robotics and Automation Letters}, 2020{\natexlab{a}}.

\bibitem[Luiten et~al.(2020{\natexlab{b}})Luiten, Osep, Dendorfer, Torr,
  Geiger, Leal-Taixe, and Leibe]{Luiten2020IJCV}
J.~Luiten, A.~Osep, P.~Dendorfer, P.~Torr, A.~Geiger, L.~Leal-Taixe, and
  B.~Leibe.
\newblock Hota: A higher order metric for evaluating multi-object tracking.
\newblock \emph{International Journal of Computer Vision (IJCV)},
  2020{\natexlab{b}}.

\bibitem[min Song et~al.(2021)min Song, chul Yoon, Yoon, Jeon, Lee, and
  Pedrycz]{song2021online}
Y.~min Song, Y.~chul Yoon, K.~Yoon, M.~Jeon, S.-W. Lee, and W.~Pedrycz.
\newblock Online multi-object tracking and segmentation with gmphd filter and
  mask-based affinity fusion, 2021.

\bibitem[Ooi et~al.(2018)Ooi, Bilodeau, Saunier, and
  Beaupr{\'e}]{ooi2018multiple}
H.-L. Ooi, G.-A. Bilodeau, N.~Saunier, and D.-A. Beaupr{\'e}.
\newblock Multiple object tracking in urban traffic scenes with a multiclass
  object detector.
\newblock In \emph{International Symposium on Visual Computing}, pages
  727--736. Springer, 2018.

\bibitem[Perreault et~al.(2021)Perreault, Bilodeau, Saunier, and
  H{\'e}ritier]{perreault2021centerpoly}
H.~Perreault, G.-A. Bilodeau, N.~Saunier, and M.~H{\'e}ritier.
\newblock Centerpoly: real-time instance segmentation using bounding polygons.
\newblock \emph{arXiv preprint arXiv:2108.08923}, 2021.

\bibitem[Qiao et~al.(2021)Qiao, Zhu, Adam, Yuille, and Chen]{vip_deeplab}
S.~Qiao, Y.~Zhu, H.~Adam, A.~Yuille, and L.-C. Chen.
\newblock Vip-deeplab: Learning visual perception with depth-aware video
  panoptic segmentation.
\newblock \emph{Proceedings of the IEEE Conference on Computer Vision and
  Pattern Recognition}, 2021.

\bibitem[Ren et~al.(2017)Ren, Chen, Liu, Sun, Pang, Yan, Tai, and
  Xu]{ren2017accurate}
J.~Ren, X.~Chen, J.~Liu, W.~Sun, J.~Pang, Q.~Yan, Y.-W. Tai, and L.~Xu.
\newblock Accurate single stage detector using recurrent rolling convolution.
\newblock In \emph{Proceedings of the IEEE conference on computer vision and
  pattern recognition}, pages 5420--5428, 2017.

\bibitem[Ren et~al.(2015)Ren, He, Girshick, and Sun]{ren2015faster}
S.~Ren, K.~He, R.~Girshick, and J.~Sun.
\newblock Faster r-cnn: Towards real-time object detection with region proposal
  networks.
\newblock \emph{Advances in neural information processing systems},
  28:\penalty0 91--99, 2015.

\bibitem[Schulter et~al.(2017)Schulter, Vernaza, Choi, and
  Chandraker]{schulter2017deep}
S.~Schulter, P.~Vernaza, W.~Choi, and M.~Chandraker.
\newblock Deep network flow for multi-object tracking.
\newblock In \emph{Proceedings of the IEEE Conference on Computer Vision and
  Pattern Recognition}, pages 6951--6960, 2017.

\bibitem[Sharma et~al.(2018)Sharma, Ansari, Murthy, and
  Krishna]{sharma2018beyond}
S.~Sharma, J.~A. Ansari, J.~K. Murthy, and K.~M. Krishna.
\newblock Beyond pixels: Leveraging geometry and shape cues for online
  multi-object tracking.
\newblock In \emph{2018 IEEE International Conference on Robotics and
  Automation (ICRA)}, pages 3508--3515. IEEE, 2018.

\bibitem[Song and Jeon(2020{\natexlab{a}})]{gmphdsaf}
Y.-M. Song and M.~Jeon.
\newblock Online multi-object tracking and segmentation with gmphd filter and
  simple affinity fusion.
\newblock \emph{arXiv preprint arXiv:2009.00100}, 2020{\natexlab{a}}.

\bibitem[Song and Jeon(2020{\natexlab{b}})]{song2020online}
Y.-m. Song and M.~Jeon.
\newblock Online multi-object tracking and segmentation with gmphd filter and
  simple affinity fusion.
\newblock \emph{arXiv preprint arXiv:2009.00100}, 2020{\natexlab{b}}.

\bibitem[Tang et~al.(2017)Tang, Andriluka, Andres, and
  Schiele]{tang2017multiple}
S.~Tang, M.~Andriluka, B.~Andres, and B.~Schiele.
\newblock Multiple people tracking by lifted multicut and person
  re-identification.
\newblock In \emph{Proceedings of the IEEE conference on computer vision and
  pattern recognition}, pages 3539--3548, 2017.

\bibitem[Voigtlaender et~al.(2019)Voigtlaender, Krause, O\u{s}ep, Luiten,
  Sekar, Geiger, and Leibe]{Voigtlaender19CVPR_MOTS}
P.~Voigtlaender, M.~Krause, A.~O\u{s}ep, J.~Luiten, B.~B.~G. Sekar, A.~Geiger,
  and B.~Leibe.
\newblock {MOTS}: Multi-object tracking and segmentation.
\newblock In \emph{CVPR}, 2019.

\bibitem[Wang et~al.(2021)Wang, Zhao, Li, Wang, Torr, and
  Bertinetto]{wang2021different}
Z.~Wang, H.~Zhao, Y.-L. Li, S.~Wang, P.~H. Torr, and L.~Bertinetto.
\newblock Do different tracking tasks require different appearance models?
\newblock \emph{arXiv preprint arXiv:2107.02156}, 2021.

\bibitem[Wojke et~al.(2017)Wojke, Bewley, and Paulus]{wojke2017simple}
N.~Wojke, A.~Bewley, and D.~Paulus.
\newblock Simple online and realtime tracking with a deep association metric.
\newblock In \emph{2017 IEEE international conference on image processing
  (ICIP)}, pages 3645--3649. IEEE, 2017.

\bibitem[Xu et~al.(2019)Xu, Cao, Zhang, and Hu]{xu2019spatial}
J.~Xu, Y.~Cao, Z.~Zhang, and H.~Hu.
\newblock Spatial-temporal relation networks for multi-object tracking.
\newblock In \emph{Proceedings of the IEEE/CVF International Conference on
  Computer Vision}, pages 3988--3998, 2019.

\bibitem[Xu et~al.(2020)Xu, Zhang, Tan, Yang, Huang, Wen, Ding, and
  Huang]{xu2020Segment}
Z.~Xu, W.~Zhang, X.~Tan, W.~Yang, H.~Huang, S.~Wen, E.~Ding, and L.~Huang.
\newblock Segment as points for efficient online multi-object tracking and
  segmentation.
\newblock In \emph{Proceedings of the European Conference on Computer Vision
  (ECCV)}, 2020.

\bibitem[Yang et~al.(2016)Yang, Choi, and Lin]{yang2016exploit}
F.~Yang, W.~Choi, and Y.~Lin.
\newblock Exploit all the layers: Fast and accurate cnn object detector with
  scale dependent pooling and cascaded rejection classifiers.
\newblock In \emph{Proceedings of the IEEE conference on computer vision and
  pattern recognition}, pages 2129--2137, 2016.

\bibitem[Yang et~al.(2020)Yang, Chang, Dang, Zheng, Sakti, Nakamura, and
  Wu]{yang2020remots}
F.~Yang, X.~Chang, C.~Dang, Z.~Zheng, S.~Sakti, S.~Nakamura, and Y.~Wu.
\newblock Remots: Self-supervised refining multi-object tracking and
  segmentation.
\newblock \emph{arXiv preprint arXiv:2007.03200}, 2020.

\bibitem[Zhou et~al.(2020)Zhou, Koltun, and
  Kr{\"a}henb{\"u}hl]{zhou2020tracking}
X.~Zhou, V.~Koltun, and P.~Kr{\"a}henb{\"u}hl.
\newblock Tracking objects as points.
\newblock In \emph{European Conference on Computer Vision}, pages 474--490.
  Springer, 2020.

\end{thebibliography}
%%%%%%%%%%%%%%%%%%%%%%%%%%%%%%%%%%%%%%%%%%%%%%%%%%%%%%%%%%%%
\end{document}